# Omni-Directional Image Generation from Single Snapshot Image

Keisuke Okubo  
*Department of Information and Communication Sciences*  
Sophia University  
Tokyo, Japan  
keisukeookubo1997@gmail.com

Takao Yamanaka  
*Department of Information and Communication Sciences*  
Sophia University  
Tokyo, Japan  
takao-y@sophia.ac.jp

*Abstract*—An omni-directional image (ODI) is the image that has a field of view covering the entire sphere around the camera. The ODIs have begun to be used in a wide range of fields such as virtual reality (VR), robotics, and social network services. Although the contents using ODI have increased, the available images and videos are still limited, compared with widespread snapshot images. A large number of ODIs are desired not only for the VR contents, but also for training deep learning models for ODI. For these purposes, a novel computer vision task to generate ODI from a single snapshot image is proposed in this paper. To tackle this problem, the conditional generative adversarial network was applied in combination with classconditioned convolution layers. With this novel task, VR images and videos will be easily created even with a smartphone camera.

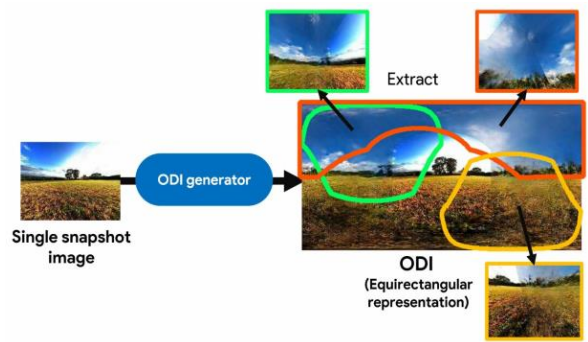

Fig. 1: Concept of ODI generation system. This system can be used to generate ODI and extract snapshot images at arbitrary locations.

## I. INTRODUCTION

Omni-directional images (ODI), the images that have a field of view covering the entire sphere around the camera, have been expected to be widely used in many applications, including virtual reality (VR), demonstrations, robotics, and social network services. However, the available images and videos of ODI are still limited compared with snapshot images taken with ordinary cameras. Therefore, it would be beneficial for these applications to synthesize ODIs from ordinary snapshot images. Furthermore, the computer-vision tasks for ODI such as object detection, semantic segmentation, scene recognition, and depth estimation have been successfully tackled with deep neural networks [1], [2]. To apply the deep learning techniques to ODI, a large number of training images of ODIs are required to achieve accurate estimation. The synthesis of ODIs from snapshot images would be useful to create such large ODI datasets.

For these purposes, a novel computer-vision task to generate ODI from a single snapshot image is proposed in this paper, as shown in Fig. 1. Since many snapshot images are widely available, a large number of ODIs can be easily created from the snapshot images. It is noted that the main purpose of the task is not to accurately reconstruct the grand truth of ODI, but to generate a natural ODI including the snapshot scene as a part of ODI. The generated ODIs can be used for developing VR applications, in addition to developing image processing methods for ODI with deep learning techniques.

Recently, many image generation and conversion methods using generative adversarial network (GAN) [3] have been developed. For example, conditional GAN (cGAN) [4] can color line drawings or generate natural images from object label maps. Most of these techniques have been applied to the images taken with ordinary cameras, but have not been applied to ODI. In this paper, cGAN was adopted for generating ODI from a single snapshot image.

In contrast to the snapshot image, ODI has a property of continuities at the edge of the image (left and right, top, bottom). To incorporate this property in cGAN, a padding method with pixel values at the other side was introduced before the discriminator in cGAN, to easily recognize fake images without the continuities. Furthermore, a class-conditioned convolution layer was developed to incorporate the class information to cGAN. Attention weights for feature-map channels in each class-conditioned convolution layer are calculated from the scene class label obtained from scene recognition of the input snapshot image, and then are applied to the feature map of the conventional convolution or deconvolution layer. All the convolution and deconvolution layers in cGAN were replaced to this class-conditioned convolution layers to realize the ODI generator conditioned on the scene class.

The contributions of this paper include:
- To propose a novel computer-vision task to generate ODI from a single snapshot image.

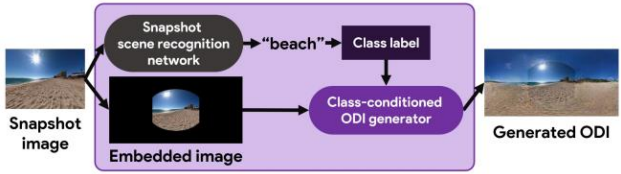

Fig. 2: Structure of class-conditioned ODI generator from single snapshot image.

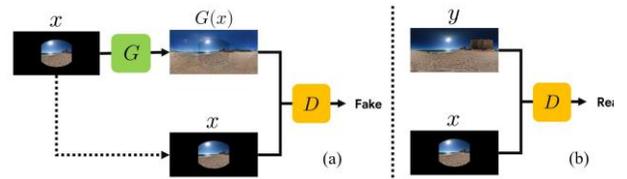

Fig. 3: Network structure of ODI generator based on cGAN. During training the discriminator ($D$), labels are set to 'Fake' and 'Real' for (a) and (b), respectively. During training the generator ($G$), the label is set to 'Real' for (a).

- To apply cGAN to generating ODI by padding generated images before the discriminator to realize the continuities of ODI.
- To develop a class-conditioned convolution layer to realize the ODI generator conditioned on the scene class.

The implementation of the proposed method will be available from https://github.com/keisuke-okb/ class-conditioned-ODI-generator-pytorch.

## II. RELATED WORKS

The most similar study to the ODI generation in this paper is the method of reconstructing a panorama image in an indoor scene from a partial observation [5]. This study has used RGBD information to predict a 3D structure and a probability distribution of semantic labels for a full 360 panoramic view. The important difference of our proposed method with this study is that the proposed ODI generation task does not aim at reconstructing the actual scene, but creating natural images with a partial snapshot image.

For omni-directional image processing, several works have attempted to apply deep convolutional neural networks (CNN) to realize classification, object detection, semantic segmentation, and depth prediction [1], [2]. They have created distorted convolutional filters depending on the locations in the equirectangular form of ODI to process the accurate shapes on the sphere. Although this type of convolution filters has not yet been applied to the methods in this paper due to the limitation of the GPU memory, they will improve the quality of generated ODI images.

Most of the recent image generation techniques are mainly based on GAN [3] composed of a generator and a discriminator. The generator is trained to generate an image that the discriminator judges as real, whereas the discriminator is trained to discriminate real and fake images. With the success of GAN, many improved and advanced methods have been proposed [6]. Among them, StackGAN [7] is an image generation method from a textual description of an animal or thing. StyleGAN [8] can output high resolution images by applying a style conversion technique. In addition, WGANgp [9] and SN-GAN [10] are advanced image generation methods by modifying the loss function, and the structure of the generator and the discriminator. cGAN [4] is an image to image translation method which converts an image to another related image. This cGAN was adopted in this paper as a method to realize ODI generation from a single snapshot image, though more recent advanced techniques can also be applied.

## III. METHODS

### A. Outline of ODI Generator

The ODI generator proposed in this paper is shown in Fig. 2. ODI is generated in the equirectangular projection from a single snapshot image by extrapolating the surrounding region of the snapshot image using cGAN-based generator. A snapshot image is embedded in the equirectangular projection of ODI to input to the generator. Although the generator could be constructed with a single cGAN-based generator without scene class information (class-independent generator), the quality of the generated ODI would be better with the scene class information. A naive method to incorporate the class information is to construct multiple cGAN-based generators for scene classes such as a generator for each class (classspecific generator), but it is not efficient. Therefore, a classconditioned convolution layer was developed to incorporate the class information with a single cGAN-based generator. To generate the scene class label, the snapshot image is first classified into a scene class with a scene recognition network (Section III-B), and then input to the cGAN-based generator with class-conditioned convolution layers (Section III-D) after embedded in the equirectangular projection (Section III-C).

### B. Scene Recognition for Snapshot Images

To generate the class label for the class-conditioned generator, a scene recognition network based on ResNet, Places365CNN [11], was used in this study. Although any scene recognition networks can be used in principle, the ResNetbased model was adopted due to its high performance in the recognition and the availability of the source code. The network was fine-tuned using the snapshot images extracted from the ODI dataset.

### C. Embedding of Snapshot Image in ODI

An input snapshot image is embedded in equirectangular projection before inputting to the ODI generator, as shown in

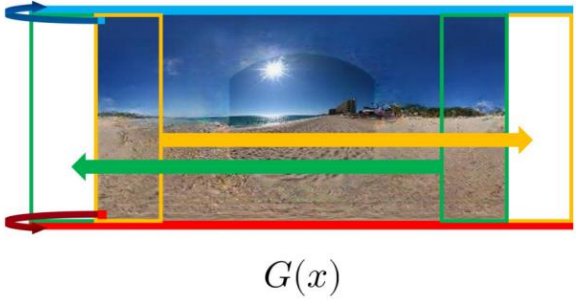

$G(x)$

Fig. 4: Padding method for 4 sides of input image for discriminator, by copying the pixel values at the opposite side for left or right edge, and by copying a pixel value at the top or bottom edge.

Fig. 2. When the camera direction is $(\theta_c, \phi_c)$ in the polar coordinate, unit vectors in the snapshot image in the 3D Euclidean coordinate system are given by the following equations.

$$X_n = (-\sin\theta_c, -\cos\theta_c, 0)$$
$$Y_n = (-\sin\phi_c \cos\theta_c, \sin\phi_c \sin\theta_c, \cos\phi_c) \quad (1)$$

With this unit vectors, the 2D coordinates in the snapshot image are transformed into the 3D coordinates, and then are represented in the polar coordinates to locate the points in the equirectangular projection. The snapshot image can be embedded using the located points in the equirectangular projection. The outside of the embedded snapshot image in ODI leaves to blank, as shown in Fig. 2. The extraction of a snapshot image from ODI can also be performed using the unit vectors, and used for creating the training dataset for the snapshot scene recognition system (Section III-B).

*D. cGAN-based ODI Generator*

The ODI generator in Fig. 2 is based on cGAN [4], which can be used for image to image translation. In our purpose, cGAN is used for translating a snapshot image embedded in equirectangular projection into ODI by extrapolating the outside of the snapshot image region. The objective of cGAN can be expressed as

$$\mathcal{L}_{cGAN}(G, D) = \mathbb{E}_{x,y}[\log D(x, y)] \\ + \mathbb{E}_{x,z}[\log(1 - D(x, G(x, z)))], \quad (2)$$

where the generator ($G$) tries to minimize this objective against an adversarial discriminator ($D$) that tries to maximize it. *x*, *y*, and *z* represent an input image, the optimum output, and a random vector, respectively. On the contrary, the objective of GAN [3] is expressed as

$$\mathcal{L}_{GAN}(G, D) = \mathbb{E}_y[\log D(y)] \\ + \mathbb{E}_{x,z}[\log(1 - D(G(x, z)))], \quad (3)$$

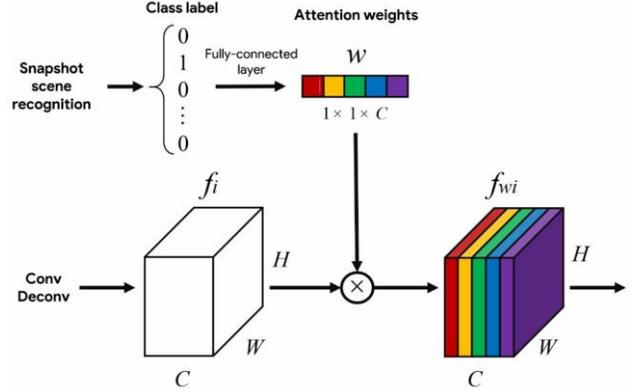

Fig. 5: Structure of class-conditioned convolution layer with attention weights in class-conditioned ODI generator model. The class-conditioned convolution layer was used for each convolution or deconvolution layer in the generator and the discriminator of cGAN.

where the difference from cGAN is that $D$ does not receive the $G$ input *x*. By incorporating the $G$ input in $D$, $G$ outputs the images paired with *x* since $D$ can easily distinguish fake images that are not paired with *x*, as shown in Fig. 3. In addition, the following reconstruction loss is also added to the objective to train cGAN.

$$\mathcal{L}_{L1}(G) = \mathbb{E}_{x,y,z}[\|y - G(x, z)\|_1] \quad (4)$$

Thus, the final objective is

$$G^* = \arg\min_G \max_D \mathcal{L}_{cGAN}(G, D) + \lambda \mathcal{L}_{L1}(G). \quad (5)$$

In contrast to the original cGAN for snapshot images, the generated ODI needs to have the property of continuities at the edges of the image. That is, the left side of ODI should be connected to the right side, whereas all the pixels at the top or bottom edge of ODI should be connected together. To realize this property, a padding method is applied before $D$ for easily distinguishing fake images without the continuities at the edges. The padding method used in this paper is shown in Fig. 4. The left and right sides of the generated image are padded with the regions at the other sides, respectively. In addition, any one pixel at the top (bottom) edge is copied to a row above (below) the image. With this padding, $G$ generates ODI with the property of the continuities at the edges, since $D$ can easily recognize fake images based on the continuities.

In addition to the padding, class-conditioned convolution layers were introduced to cGAN. Since multiple ODI generators for scene classes are inefficient, convolution layers in a single ODI generator are conditioned on the scene class label obtained from scene recognition for a snapshot image by weighing the channels of convolution filters depending on the scene label. This attention weights are calculated from the scene label using a fully-connected layer, and then are applied to the feature map from the convolution layer, as shown in Fig. 5. All the convolution/deconvolution layers were replaced by this class-conditioned convolution/deconvolution layers, whose attention weights were independent over all the layers.

IV. EXPERIMENTAL SETUP

*A. Dataset*

The dataset used in the experiments is SUN360 [12]. This dataset includes images in 30 outdoor scenes excluding 'others'. The numbers of images in the dataset are imbalanced among classes. Since the 6 scene classes include less than 10 images, these classes were excluded from the experiments. The numbers of images in the remaining 24 scene classes range from 11 to 690. Since the dataset is still imbalanced among the classes, the performance of the network may be affected by the imbalance among the classes. Therefore, the performance for training networks with all images was compared to the performance with limiting the maximum number of images for each class.

In the dataset, the two sizes of ODIs are available: 1024 × 512 and 256 × 128. The ODIs in 1024 × 512 were used with downsampling to 512 × 256 to train the network efficiently. The dataset was randomly divided into 75% images for the training dataset and 25% images for the test dataset. The ODI generator and the snapshot scene recognition network were trained using the training dataset, and then evaluated with the test dataset.

*B. Learning ODI Generator*

The ODI generator based on cGAN used in this paper was adapted from the pix2pix implementation in PyTorch [13]. In the model, the generator was composed of U-Net [14] (8 convolution layers in the encoder and 8 deconvolution layers in the decoder). The discriminator was realized with the network of 5 convolution layers. The parameters were set to the default values in this implementation. The convolution and deconvolution layers were set to 64, 96, and 128 channels to compare the expressive power of the network. The width of the padding region was set to 50 pixels for both sides of the generated image in Fig. 4.

To train the ODI generator based on cGAN, the snapshot images were prepared by extracting them from ODIs in the outdoor scene dataset. The relationship between the view angle of the camera for the extraction and the size of the extracted snapshot image is given by the following equations:

$$\theta_a = 2\arctan(W_1/2L)$$
$$\phi_a = 2\arctan(H_1/2L), \quad (6)$$

where $\theta_a$ and $\phi_a$ are the horizontal and vertical view angles, while $W_1$ and $H_1$ are the width and height of the extracted image. $L$ is the distance from the camera to the image plane. In the experiments, they were set to the following values: $L = 100$, $W_1 = 400$, $H_1 = 300$. The camera direction, $\theta_c$ and $\phi_c$ in Eq. (1), was set to 0 degree for both latitude and longitude. One snapshot image was extracted from one ODI in the dataset. Using the pairs of the original ODIs and the extracted snapshot images, the ODI generator was trained for all classes with the class labels.

*C. Learning Scene Recognition for Snapshot Image*

The scene recognition network for snapshot images was adapted from Places365-CNN [11]. Among several implementations with various base CNN models, the model based on ResNet [15] with 18 layers (ResNet18 model) was used in the experiments.

The network was fine-tuned using the snapshot images extracted from ODIs in the training dataset. To extract the snapshot images, the camera directions were set to 0, 60, 120, 180, 240, and 300 degrees in the longitude with 0 degree in the latitude (6 camera directions). The parameters for the fine tuning were set to the default values in the implementation.

*D. Evaluation Measure for ODI Generator*

An important problem for the image generation task is to design a performance measure to evaluate the generated images. The ODIs generated using the proposed methods should be categorized into the same scene class as the original ODI from which the input snapshot image was extracted. In the experiments, the following two methods were used for the evaluation.

The first is to use a scene recognition network for ODI in the equirectangular projection. It is noted that this network is different from the scene recognition network for snapshot images in Section IV-C, although the same architecture, ResNet18 model in Places365-CNN, was also used for the ODI scene recognition network. To train the network, all images in the ODI dataset including the training and test datasets were used. The inputs of the network were ODIs in the equirectangular projection to classify the input ODI into a scene class from the 24 classes.

The second is to use a scene recognition network for several extracted snapshot images from a generated ODI. In this method, the scene recognition network in Section IV-C can be used as is. In the experiments, snapshot images were extracted from a generated ODI in 10 horizontal directions.

Using these evaluation methods, the proposed ODI generation methods were evaluated by calculating the number of images that were categorized into the same class as the original ODI. Since the random vectors were input to the ODI generator to generate a variety of images, the performance of the ODI generators was evaluated with the average recognition rate of the scene recognition for ODI over 5 repetitions.

In addition to these metrics, a standard GAN evaluation metric, Frechet Inception Distance (FID) [16], was also calculated for the generated ODIs with a small modification, that is the use of the ODI scene recognition model above instead of the Inception network in FID.

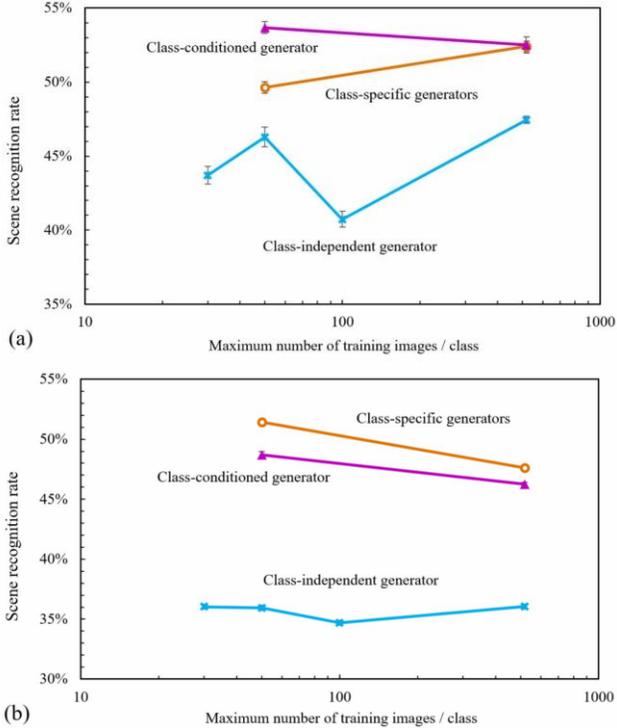

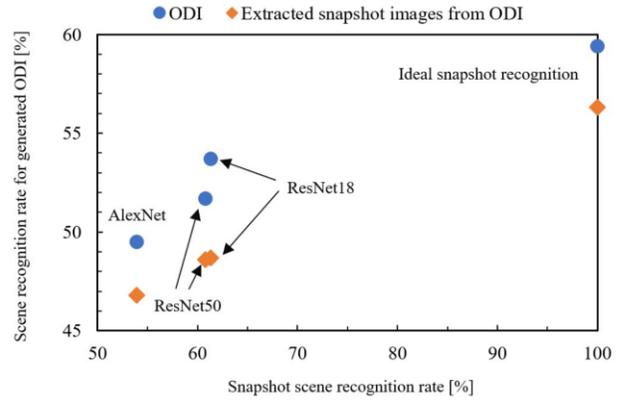

Fig. 8: Comparison of scene recognition rates for generated ODI against scene recognition rates for input snapshot image.

of the image, and the average value, respectively. The standard deviation of the bottom side $\sigma_B$ was also calculated. For the left and right sides, the root mean square of the difference between the left and right side pixels was calculated as

$$\sigma_{LR} = \sqrt{\frac{\Sigma_y (v_L(y) - v_R(y))^2}{N_H}}, \qquad (8)$$

where $y$, $v_L$, $v_R$, $N_H$ are the pixel position in the height direction, the pixel value at the left side, the pixel value at the right side, and the height of the image, respectively. In both equations (7) and (8), the smaller value represents the better continuities in the generated ODI in the equirectangular projection.

## V. EXPERIMENTAL RESULTS

### A. Comparison of Proposed Method with Baseline Methods

The ODI generator with class-conditioned convolution layers (128ch) was compared with two baselines: a single ODI generator without the class-conditioned convolution layers (class-independent generator), and an ODI generator for each class (class-specific generator). In the class-specific generator, an input snapshot image was classified into a scene category, and then was input to the ODI generator trained for the specific scene class. Although the quality of the generated ODI was improved by the class-specific generator, it is inefficient since many generators need to be independently trained for all the scene classes.

The experimental results are shown in Fig. 6, where the average of recognition rates over all the scene classes is plotted against the maximum number of training images for each class. The training images for each class was limited since the dataset was imbalanced as explained in Section IV-A. It can be seen from the figure that the performance of the class-conditioned ODI generator was much better than the class-independent ODI generator, and was comparable to the class-specific ODI generator which required much higher computational cost.

Fig. 6: Comparison of scene recognition rates for generated ODI. (a) ODI in equirectangular projection. (b) Snapshot images extracted from generated ODI.

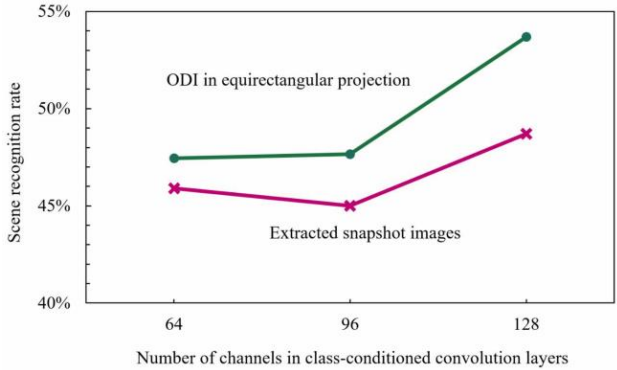

Fig. 7: Comparison of scene recognition rates in different number of channels for class-conditioned convolution layers.

### E. Evaluation Measure for Padding

As shown in Section III-D, the image with padding is inputted to the discriminator in order to reflect the property of continuities. The following equations were used to evaluate the continuities of generated ODI in the equirectangular projection with and without the padding. At the top and bottom sides, the standard deviations of the pixel values were calculated as

$$\sigma_T = \sqrt{\frac{\Sigma_x (v_T(x) - \overline{v_T(x)})^2}{N_W}}, \qquad (7)$$

where $x$, $v_T$, $N_W$, $v_T(x)$ are the pixel position in the horizontal direction, the pixel value at the top side of the image, the width

TABLE I: Comparison of class-conditioned ODI generator to baseline methods (class-independent generator and class-specific generators). The ODI generation was repeated 5 times for each snapshot image input. The values outside and inside the parentheses indicate the scene recognition rates for ODI in equirectangular projection and for extracted snapshot images from ODI, respectively.

| Class | # of training images | # of test images | Original ODI | Baselines | | Proposed |
|---|---|---|---|---|---|---|
| | | | | Class-independent generator | Class-specific generators | Class-conditioned generator |
| Arena | 15 | 5 | 60.0% (60.0%) | 0.0% (24.4%) | 4.0% (43.2%) | 20.0% (32.0%) |
| Balcony | 50 | 17 | 100.0% (74.1%) | 47.1% (19.5%) | 68.2% (29.3%) | 65.9% (42.8%) |
| Beach | 147 | 49 | 69.4% (60.6%) | 67.3% (57.4%) | 73.9% (66.3%) | 71.4% (60.0%) |
| Boat | 55 | 18 | 83.3% (66.1%) | 47.8% (46.3%) | 62.2% (70.3%) | 67.8% (50.9%) |
| Bridge | 48 | 16 | 87.5% (63.8%) | 17.5% (15.5%) | 37.5% (38.3%) | 32.5% (24.3%) |
| Cemetery | 28 | 9 | 88.9% (53.3%) | 77.8% (18.2%) | 44.4% (22.7%) | 62.2% (28.4%) |
| Coast | 58 | 20 | 90.0% (60.0%) | 77.0% (26.2%) | 63.0% (41.4%) | 74.0% (41.7%) |
| Desert | 35 | 12 | 100.0% (78.3%) | 56.7% (30.8%) | 83.3% (70.5%) | 86.7% (64.5%) |
| Field | 248 | 82 | 72.0% (71.8%) | 67.1% (63.7%) | 69.5% (66.1%) | 65.9% (66.6%) |
| Forest | 179 | 60 | 98.3% (85.3%) | 88.3% (73.4%) | 96.0% (93.5%) | 94.0% (87.5%) |
| Highway | 10 | 3 | 100.0% (40.0%) | 20.0% (12.7%) | 6.7% (6.0%) | 0.0% (31.3%) |
| Jetty | 28 | 9 | 100.0% (68.9%) | 44.4% (28.7%) | 88.9% (64.0%) | 86.7% (62.9%) |
| Lawn | 75 | 25 | 80.0% (80.8%) | 61.6% (68.9%) | 56.0% (67.0%) | 62.4% (75.0%) |
| Mountain | 193 | 64 | 95.3% (70.8%) | 80.9% (61.3%) | 87.5% (77.5%) | 79.7% (68.6%) |
| Park | 120 | 40 | 57.5% (55.0%) | 32.0% (39.9%) | 29.0% (31.5%) | 34.0% (51.6%) |
| Parking lot | 50 | 16 | 56.3% (66.9%) | 5.0% (24.5%) | 13.8% (30.3%) | 5.0% (32.0%) |
| Patio | 13 | 4 | 75.0% (45.0%) | 45.0% (26.5%) | 40.0% (24.5%) | 40.0% (31.0%) |
| Plaza courtyard | 518 | 172 | 63.4% (57.3%) | 37.4% (32.3%) | 53.1% (50.8%) | 42.8% (35.5%) |
| Ruin | 84 | 28 | 85.7% (83.2%) | 51.4% (38.7%) | 58.6% (63.6%) | 59.3% (78.7%) |
| Sports field | 8 | 3 | 100.0% (26.7%) | 46.7% (14.0%) | 46.7% (33.3%) | 80.0% (30.7%) |
| Street | 484 | 162 | 63.6% (48.5%) | 39.6% (38.4%) | 43.7% (45.4%) | 21.9% (28.5%) |
| Swimming pool | 27 | 9 | 100.0% (64.4%) | 80.0% (56.7%) | 66.7% (51.3%) | 80.0% (68.0%) |
| Train station or track | 30 | 10 | 60.0% (60.0%) | 8.0% (19.0%) | 46.0% (30.0%) | 10.0% (27.0%) |
| Wharf | 83 | 28 | 71.4% (52.9%) | 40.0% (28.3%) | 19.3% (25.9%) | 46.4% (49.2%) |
| Average | | | 81.6% (62.2%) | 47.4% (36.1%) | 52.4% (47.6%) | 53.7% (48.7%) |
| FID | | | 0.00 | 24.7 | 22.1 | 20.8 |

TABLE II: Evaluation of continuities with or without padding. In all cases, the average of 24 classes is shown.

| Side | Original ODI | Without padding | | With padding | |
|---|---|---|---|---|---|
| | | Class-conditioned generator | Class-specific generators | Class-conditioned generator | Class-specific generators |
| Top ($\sigma_T$) | 5.8562 | 9.0413 | 11.5793 | **5.8304** | 7.8358 |
| Bottom ($\sigma_B$) | 12.7523 | 12.7796 | 15.5457 | **8.8667** | 14.3549 |
| Left/Right ($\sigma_{LR}$) | 19.3376 | 39.2881 | 44.4291 | **28.8816** | 34.0271 |

Furthermore, the recognition rate for each class in the proposed method and the baseline methods is shown in Table I. In each method, the maximum number of training data per class was set to that in the best performance in Fig. 6 (a). For the reference, the scene recognition rate was also shown for the original ODIs, which was the upper limit of the performance. As can be seen from the table, the average of the recognition rates for the proposed method was higher than the baseline methods, although the recognition rates varied among the scene classes. In addition, FID with the ODI scene recognition model was calculated for the proposed and baseline methods. The proposed method was also better than the baseline methods in this metric, as shown in the bottom row in Table I.

*B. Dependence on Number of Channels*

To compare the expressive power of the network in the class-conditioned generator, the convolution and deconvolution layers were set to 64, 96, and 128 channels. The comparison of the scene recognition rates in different number of channels for the class-conditioned convolution layers is shown in Fig. 7. The scene recognition rates for 128 channels were greatly improved, although the scene recognition rates for 64 and 96 channels were almost the same. This tendency was similar both in ODI in the equirectangular projection and in the extracted snapshot images.

*C. Dependence on Snapshot Scene Recognition Performance*

To investigate the dependence of ODI generator performance on the scene recognition for the input snapshot images, the scene recognition rates for the generated ODIs in the proposed method were plotted against the input snapshot scene recognition rates in Fig. 8. In addition to the ResNet18, several scene recognition networks were constructed for the snapshot scene recognition model, including Alexnet and ResNet50. The ideal scene recognition was also included, that is the case of perfect scene recognition for input snapshot image. As can be seen from Fig. 8, the ODI generator performance depended on the scene recognition for the input snapshot images.

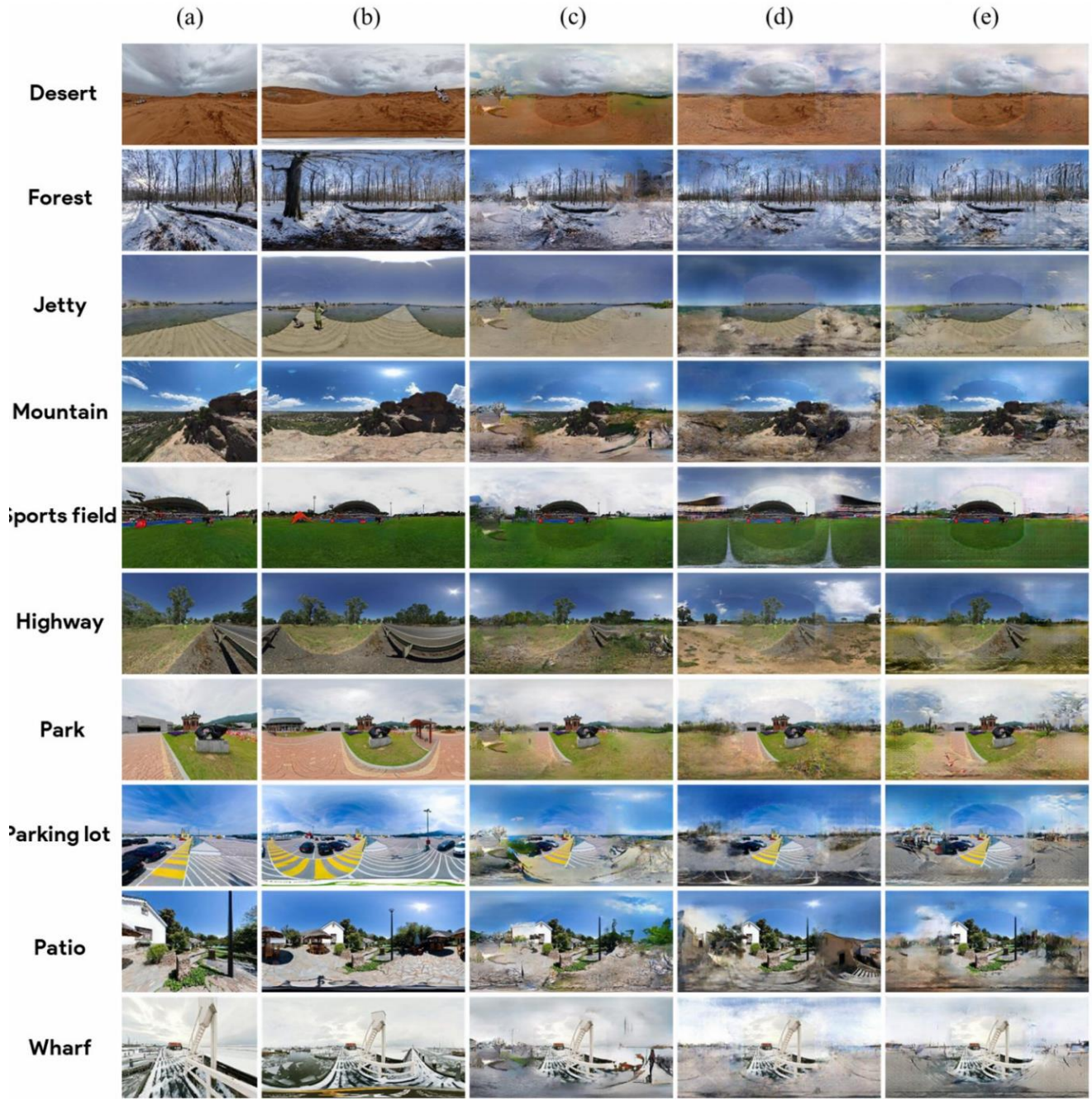

Fig. 9: Sample images of generated ODI. From the left to right, (a) input snapshot image, (b) original ODI, (c) class-independent ODI generator (baseline), (d) class-specific ODI generator (baseline), (e) class-conditioned ODI generator (proposed). The top and bottom 5 scenes are the scenes with high and low recognition rates in the evaluation.

*D. Evaluation of Padding*

To compare the continuities with or without the padding, the class-conditioned generator and the class-specific generators were trained both without and with the padding. The continuity indexes defined in Eqs. 7 and 8 were calculated for the top, bottom, and left/right sides of the generated ODI in the equirecgtangular projection. The averages over all the test images are shown in Table II. It can be seen from the results that the continuities were improved by training with padding.

*E. Generated Sample Images*

The sample images generated by the proposed ODI generator are shown in Fig. 9, in addition to the sample images by the baseline methods. As can be seen from the sample images, the natural scenes such as desert and forest can be represented in ODI, although the difference between the embedded region of the input snapshot image and its outside was still large. On the other hand, the scenes including man-made things such as a building and parking lots were difficult to be generated in all generators. From the sample images, the class-condiotiond generator (e), and the classspecific generators (d) were slightly better than the classindependent generator (c). The network architecture for the generator and the discriminator should be further investigated to improve the quality of the generated images.

*F. Computational Cost in Training and Inference*

In the proposed method (class-conditioned ODI generator) with 128 channels, it took 0.13 seconds in GPU (NVIDIA GeForce GTX1080Ti) for generating ODI from a snapshot image, compared with 0.09 seconds for the class-independent generator and 0.17 seconds for the class-specific generator. For training of the generators, it took 24 hours in the classconditioned generator, and 40 hours in the class-specific generators. Furthermore, the amount of network weights in the proposed method (800 MB) was much smaller than the amount of weights in the class-specific generators (5.0 GB). Thus, the proposed method saved the computational cost in both training and inference, in addition to the network size, compared with the class-specific generators.

## VI. CONCLUSIONS

The novel computer-vision task was proposed to generate an omni-directional image from a single snapshot image. To solve this task, cGAN-based ODI generator was developed using padding for improving the continuities in the generated ODI, and using class-conditioned convolution layers for incorporating scene class information. It was confirmed in the experiments that the performance on the proposed method of the class-conditioned generator was as high as the baseline method of the class-specific generators with lower computational cost, although still more work to improve the quality of the generated images is needed.

In the future work, it will be possible to improve the generator and the discriminator networks by incorporating the recent advance in GAN-based image generation techniques. Especially, the resolution of the generated images can be improved by applying these techniques. Futhermore, the convolution filters designed for omni-directional images can be also applied in the networks. This will remove the distortion of the generated images.